\documentclass[pmlr]{jmlr}

\RequirePackage{graphicx}
 \usepackage{booktabs}
\usepackage{longtable}
\makeatletter
\def\set@curr@file#1{\def\@curr@file{#1}} 
\makeatother
\usepackage[load-configurations=version-1]{siunitx} 

\theorembodyfont{\upshape}
\theoremheaderfont{\scshape}
\theorempostheader{:}
\theoremsep{\newline}

\title[Heuristic Step Planning for Learning Dynamic Bipedal Locomotion]{Heuristic Step Planning for Learning Dynamic Bipedal Locomotion: A Comparative Study of Model-Based and Model-Free Approaches}

\author{\Name{William Suliman}
       \Email{suliman.vil@phystech.edu}\\ 
       \addr Moscow Institute of Physics and Technology \\
	141701 Dolgoprudny, Russia 
       \AND
       \Name{Ekaterina Chaikovskaia}
       \Email{dorzhieva.em@mipt.ru}\\ 
       \addr Moscow Institute of Physics and Technology\\
       141701 Dolgoprudny, Russia 
       \AND
       \Name{Egor Davydenko}
       \Email{davydenko.ev@mipt.ru}\\ 
       \addr Moscow Institute of Physics and Technology\\
       141701 Dolgoprudny, Russia 
       \AND
       \Name{Roman Gorbachev}
       \Email{gorbachev.ra@mipt.ru}\\ 
       \addr Moscow Institute of Physics and Technology\\
       141701 Dolgoprudny, Russia} 

\begin{document}

\maketitle

\begin{abstract}

This work presents an extended framework for learning-based bipedal locomotion that incorporates a heuristic step-planning strategy guided by desired torso velocity tracking. The framework enables precise interaction between a humanoid robot and its environment, supporting tasks such as crossing gaps and accurately approaching target objects. Unlike approaches based on full or simplified dynamics, the proposed method avoids complex step planners and analytical models. Step planning is primarily driven by heuristic commands, while a Raibert-type controller modulates the foot placement length based on the error between desired and actual torso velocity. We compare our method with a model-based step-planning approach—the Linear Inverted Pendulum Model (LIPM) controller. Experimental results demonstrate that our approach attains comparable or superior accuracy in maintaining target velocity (up to 80\%), significantly greater robustness on uneven terrain (over 50\% improvement), and improved energy efficiency. These results suggest that incorporating complex analytical, model-based components into the training architecture may be unnecessary for achieving stable and robust bipedal walking, even in unstructured environments.

\end{abstract}

\section{Introduction}

Bipedal humanoid robots offer significant advantages when operating in environments designed for humans—particularly in areas where wheeled or other mobile platforms are limited. At the same time, the unique bipedal design of humanoid robots poses an engineering challenge: unlike statically stable mechanisms, these robots must constantly maintain dynamic balance and control numerous degrees of freedom. Developing a stable and energy-efficient control system for bipedal locomotion, capable of adapting to various terrains, remains one of the key challenges in robotics.

In many studies \citep{reher2021inverse,gong2019feedback}, the high-level control signal provided to the walking controller is the desired torso velocity—analogous to the control of wheeled platforms. This approach is effective on flat surfaces but lacks sufficient precision in complex terrain or constrained environments. An alternative is control through desired foot positions, which provides more reliable interaction with the environment but requires integration of a step planner.


This work aims to develop a control system for a bipedal walking robot that employs a hybrid task formulation: both desired foot positions and the target torso velocity are provided as high-level control signals to the walking controller. This allows maintaining flexible robot locomotion in open spaces while ensuring precise motion in complex environments, such as stairs or uneven surfaces. The next section presents a review of existing solutions, motivating the choice of tools and architecture for the proposed approach.

\subsection{Approaches to humanoid robot control systems}

Currently, there are two main methods for controlling bipedal walking robots. These methods form the basis of most motion control approaches, with numerous hybrid strategies lying between them. The first, more traditional approach is based on classical control methods. It involves generating reference trajectories for the robot's joints or links—either in real-time or offline. These trajectories can be obtained using full-body dynamic models, such as the Hybrid Zero Dynamics (HZD) method \citep{reher2021inverse,gong2019feedback}, or simplified models, such as the Linear Inverted Pendulum Model (LIPM) \citep{feng2016robust} and the Angular Momentum Linear Inverted Pendulum (ALIP) \citep{gong2020angular}. Once the reference trajectories are generated, the next step is to ensure accurate tracking by the robot's joints or links. Various controllers are used for this purpose, including: PD controllers \citep{gong2019feedback}, inverse dynamics controllers such as Task-Space Inverse Dynamics (TSID) \citep{delprete2016torque}, and ID-QP controllers \citep{shen2022locomotion}. Trajectory generation can be considered high-level control, while tracking represents low-level control.

The main advantage of classical approaches is their predictability—the robot's behavior can be precisely modeled in advance. However, a significant drawback is their rigid dependence on predefined trajectories, which limits the robot's adaptability in real-world conditions. Another limitation is the need to work with complex mathematical models, which even after simplification may negatively affect key performance metrics such as energy efficiency.

The second approach, closely linked to the growth of modern computational power, is model-free learning, in particular Reinforcement Learning (RL). RL has shown high potential in tasks requiring adaptability and robustness, as it allows robots to learn effective control strategies through trial-and-error without precise system modeling. This approach simplifies the process of control algorithm synthesis, but it may lead to unintuitive or physically unrealistic behavior due to the agent’s freedom in exploring the entire action space. Despite being relatively new compared to classical methods, RL-based controllers have recently made significant progress, with several researchers \citep{siekmann2021sim,liao2024berkeley} successfully implementing them on real bipedal robots.

Hybrid methods are also emerging, aiming to combine the strengths of both strategies. These approaches constrain the flexibility of RL by embedding knowledge from classical control. For instance, the HZD framework can be employed to construct a gait library of reference trajectories, while reinforcement learning is used to design a low-level controller that tracks these trajectories \citep{chaikovskaya2023benchmarking,li2024reinforcement}.

\subsection{Step Planning in Locomotion Tasks}
Reinforcement learning (RL) methods have shown significant potential for achieving dynamic locomotion in humanoid robots. However, many existing RL-based controllers do not explicitly consider foot placement constraints \citep{siekmann2021sim,margolis2023walk,liao2024berkeley}, allowing the control system to freely adjust foot positions to maintain balance. While this approach simplifies the learning problem, it limits applicability in real-world scenarios that require precise and careful foot placement, such as walking over uneven or discontinuous surfaces. Therefore, explicitly incorporating foot placement constraints into the control system is a key requirement for ensuring stable and reliable locomotion in complex and structured environments.

In this work, we propose a novel locomotion control architecture that tracks not only commanded linear and angular velocities of the robot’s body but also desired foot positions. The approach is based on a heuristic step planner that generates target foot placement positions. These targets are fed into a policy obtained via reinforcement learning methods. This structure allows for adaptive adjustment of foot placement in response to environmental changes—for example, when avoiding obstacles or bridging gaps in the support surface.

Recent years have seen several studies focused on integrating step planning into RL-based architectures for bipedal locomotion control. For instance, in \citep{wang2023learning}, the RL policy is closely coupled with a foot placement planning module, enabling determination of feasible foot positions under complex environmental conditions. Although such architectures can enhance adaptability, they also substantially increase the complexity of the control system and computational cost. Moreover, they typically require accurate estimation of global foot positions to reliably track step placement commands at each control cycle.

Our proposed approach preserves the simplicity of classical velocity-tracking control—the robot follows a nominal motion trajectory—while augmenting it with the capability to adjust foot positions when necessary. This decomposed structure promotes modularity and computational efficiency, thereby improving robustness across diverse locomotion scenarios.

A similar strategy was presented in \citep{lee2024integrating}, where target foot positions are determined using a linear inverted pendulum model (LIPM). However, we argue that the use of classical model-based planners such as LIPM is not strictly necessary and, in fact, may introduce unnecessary complexity without significant performance gain. To test this hypothesis, we conduct a comparative analysis of two step-planning-based locomotion strategies: (i) the heuristic method proposed in this work (hereafter referred to as the Linear Step Planning, or LS, approach); and (ii) the LIPM-based approach from \citep{lee2024integrating}. Through this study, we aim to address a key question in RL-based bipedal locomotion: \textbf{\textit{To what extent is knowledge derived from mathematical models necessary to guide the learning process? Moreover, how does the use of such models affect the behavior, efficiency, and generalization capability of the resulting control policies?}}

\begin{figure*}[!t]
\begin{center}
\includegraphics[width=\textwidth]{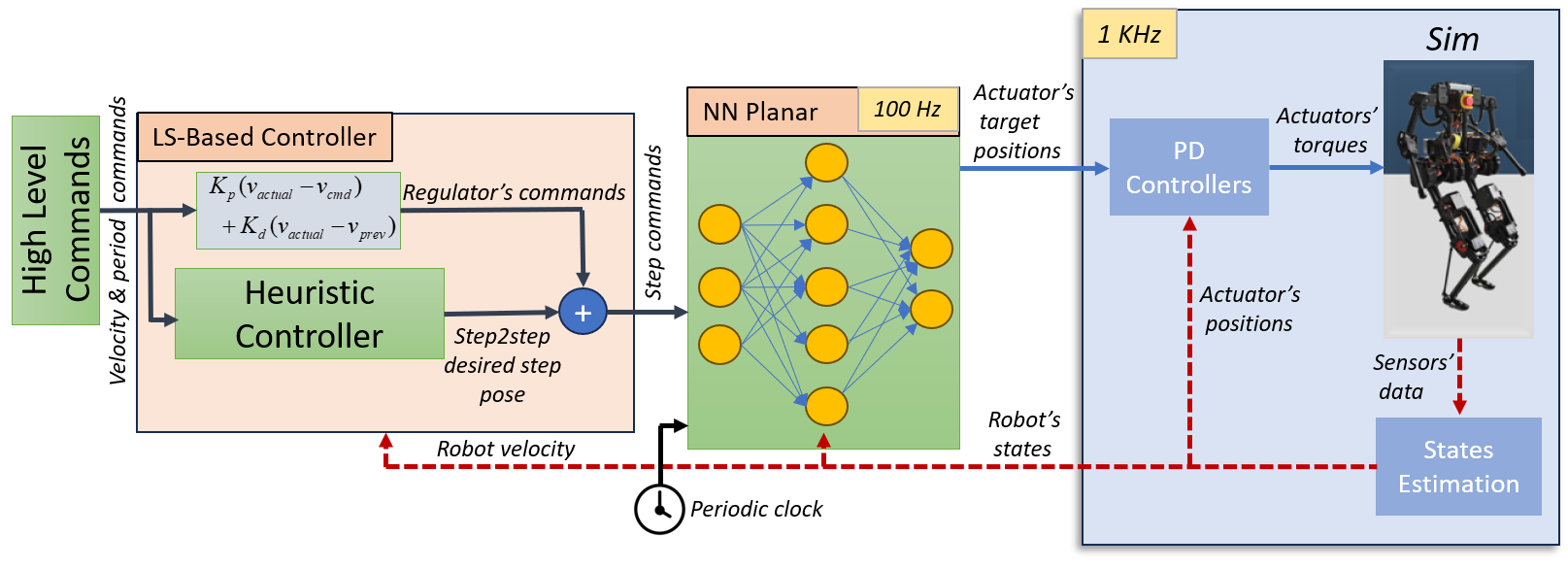}
\end{center}
\caption{An overview of the structure and flow of the proposed learning-based framework. The proposed method employs a heuristic step planner to determine step locations, incorporating Raibert-like regulators to accurately track the desired velocity.}
\label{method}
\end{figure*}

This work contributes in two main directions: (1) \textbf{Integration of heuristic step planning into the learning architecture}. We propose a simple step planning structure that does not require complex dynamic models (Fig.~\ref{method}). Unlike previous approaches where target foot positions were considered only in the RL reward function or not defined at all \citep{siekmann2021sim, margolis2023walk}, our approach employs a feedback mechanism that compares actual and target foot placements during policy execution (see Fig. 1). (2) \textbf{Comparative analysis of model-based methods}. The proposed approach is systematically compared with a simplified dynamics model (LIPM) to assess the effect of model-derived information on learning performance, locomotion stability, motion quality, and the generalization capability of the learned policy.

\begin{figure}[t]
\begin{center}
\includegraphics[width=0.5\columnwidth]{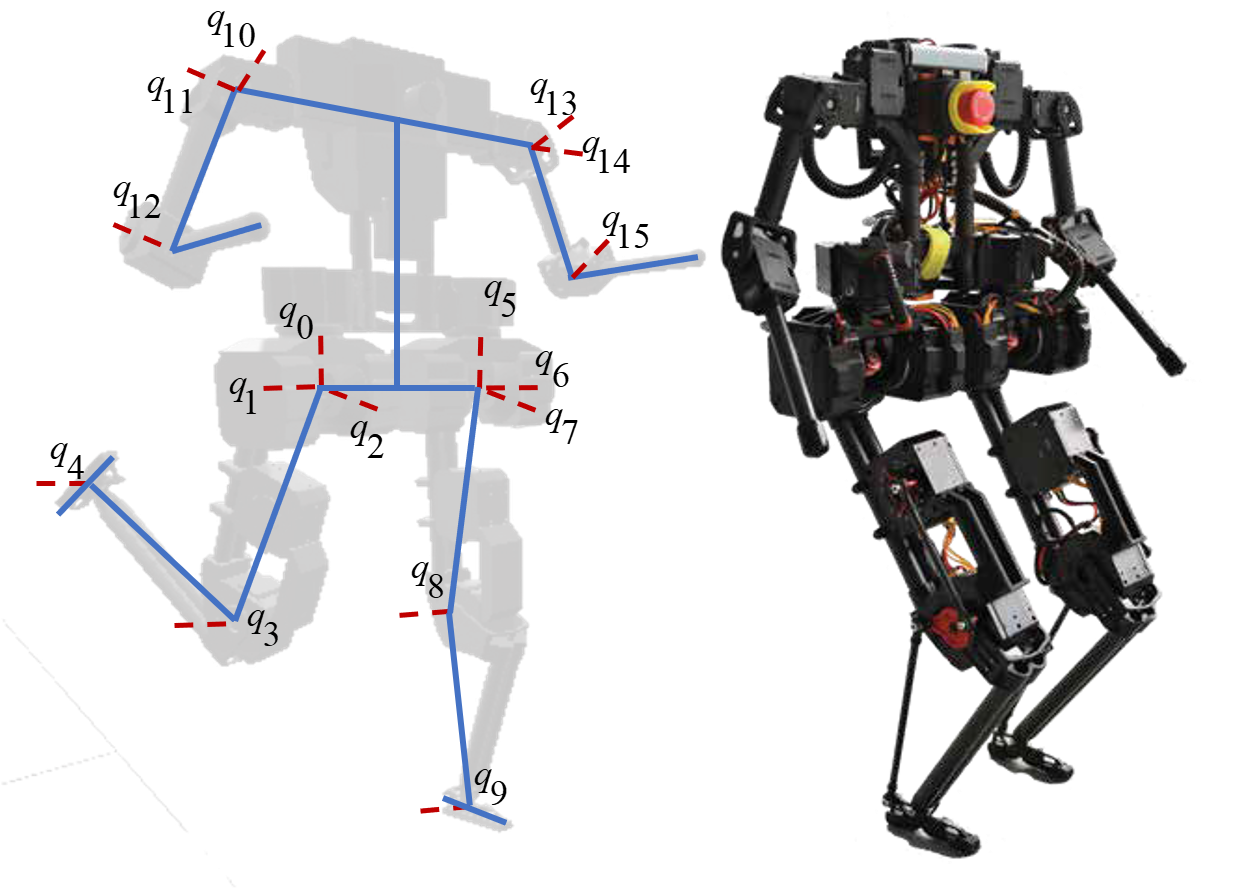}
\end{center}
\caption{BRUCE and its general morphology.}
\label{morphology}
\end{figure}

The proposed approach was validated through simulations and sim-to-sim testing on the BRUCE (Bipedal Robot Unit with Compliance Enhanced) \citep{liu2022design}, demonstrating its promise for improving robotic locomotion in challenging environments. BRUCE is a kid-size humanoid robot (Fig.~\ref{morphology}), originally developed at RoMeLa in joint effort with Westwood Robotics. It stands about 70 cm tall and weighs about 4.8 kg.

\section{Preliminaries and problem formulation}
\label{prelim}

In this section, we present the method used to build the heuristic step planner. The probosed controller is a simple high-level planner controller that calculates step length using straightforward relations, as follows:
{
\[
x_{\text{step}} = 
\begin{cases}
v_x \cdot T_d, & -l_{\text{max}} < v_x \cdot T_d < l_{\text{max}} \\
l_{\text{max}}, & v_x \cdot T_d \geq l_{\text{max}} \\
-l_{\text{max}}, & v_x \cdot T_d \leq -l_{\text{max}}
\end{cases}
\]
}
and
{
\[
y_{\text{step}} = 
\begin{cases}
v_y \cdot T_d, & 
\begin{cases}
v_y \geq 0  \text{ and left swing phase} \\
v_y \leq 0  \text{ and right swing phase}
\end{cases} \\
-\text{sign}(v_y) \cdot w_{\text{min}}, & \text{otherwise}
\end{cases}
\] }

\noindent where \( v_{x,y} \) is the robot commanded velocity, \( T_d \) is the desired step duration, \( l_{\text{max}} \) is the maximum step length, and \( w_{\text{min}} \) is the minimum step width.

A Raibert-like regulator is used to encourage the policy to accurately track a target walking velocity by adding an offset to the foot strike location \citep{raibert1984balance}:
{\small
\begin{equation}
\Delta(x,y) = K_p \left( v_k^a - v_k^d \right) + K_d \left( v_k^a - v_{k-1}^a \right),    
\end{equation}}
\noindent where \( v_k^a \) and \( v_k^d \) are the actual and desired velocity at step \( k \).

\section{RL problem formulation}
\label{RL_problem_formulation}
In an attempt to address the research question of this paper, we formulate two RL training frameworks. The first structure is shown in Fig.~\ref{method} and uses the equations presented in Section~\ref{prelim}. In addition to this structure, we also implemented the method presented in \citep{lee2024integrating}.

The control policy is a fully connected neural network with 3 hidden layers, each layer with 256 nodes. The inputs to the policy are the robot states, user velocity and step duration commands, and step commands derived from the adopted high-level controllers.

We train our policy in the Isaac Gym simulation engine on flat terrain using the PPO algorithm \citep{schulman2017ppo}. The state space, action space, and reward formulation for the RL problems are prepared as in \citep{lee2024integrating}, with the weights adjusted to fit our model as presented in Table~\ref{table_reward_weights}. The Raibert-like regulator values used for the LS-based method are
{\small \( K_p^{rb} = [0.3, 0.3] \)} and {\small \( K_d^{rb} = [0.1, 0.1] \)}.

\begin{table}[t]
\caption{Reward function weights}
\label{table_reward_weights}
\begin{center}
\begin{tabular}{llll}
\multicolumn{1}{c}{\bf Reward Term} & \multicolumn{1}{c}{\bf Weight} & \multicolumn{1}{c}{\bf Reward Term} & \multicolumn{1}{c}{\bf Weight} \\ \hline \\
Joint torques          & $1 \times 10^{-4}$  & Action smoothness 2      & 0.005 \\
Joint torque limits    & 0.01               & Hip joint regularization & 1 \\
Joint velocity        & $1 \times 10^{-3}$  & Base roll-pitch velocity & 0.01 \\
Joint limits          & 10                 & Base height              & 3 \\
Action smoothness 1   & 0.1                & Base tilting             & 1 \\
Base heading          & 3                  & Velocity tracking        & 4 \\
Contact schedule      & 9                  &                          & \\
\end{tabular}
\end{center}
\end{table}
\section{Experiment results}
\label{experiment_results}

\begin{figure}[ht]
\begin{center}
\includegraphics[width=0.8\columnwidth]{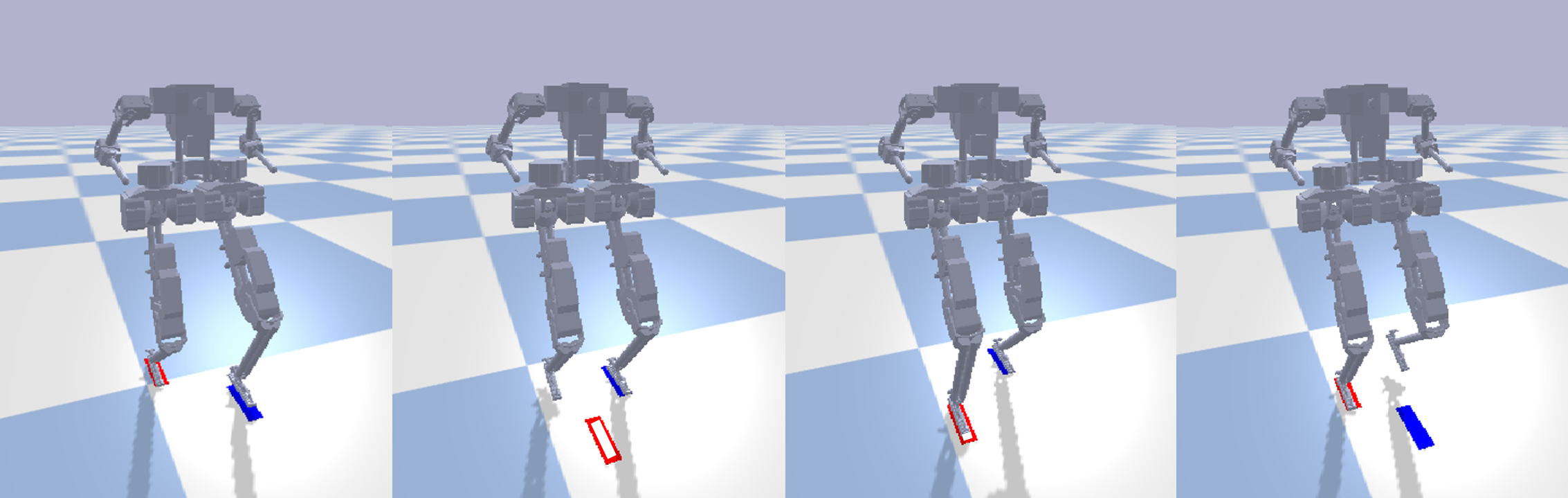}
\end{center}
\caption{Time series motion tiles for simulated walking using the LS-based approach in the PyBullet environment at \( v_x = 0.6 \, \text{m/s} \) and step duration 0.25s. Red area: right step position command, blue area: left step position command.}
\label{timeseries}
\end{figure}

We present simulation results using BRUCE and perform a sim-to-sim transfer of the control policy to the PyBullet environment to evaluate the effectiveness of these approaches.  Fig.~\ref{timeseries} shows time series motion tiles for the robot in the PyBullet environment at a forward velocity of 0.6 m/s using the LS-based approach. 

First, the simulation results are presented, followed by a detailed discussion.

\subsection{Simulation Results}
\subsubsection{Velocity tracking performance}
\begin{figure}[ht]
\begin{center}
\includegraphics[width=0.8\columnwidth]{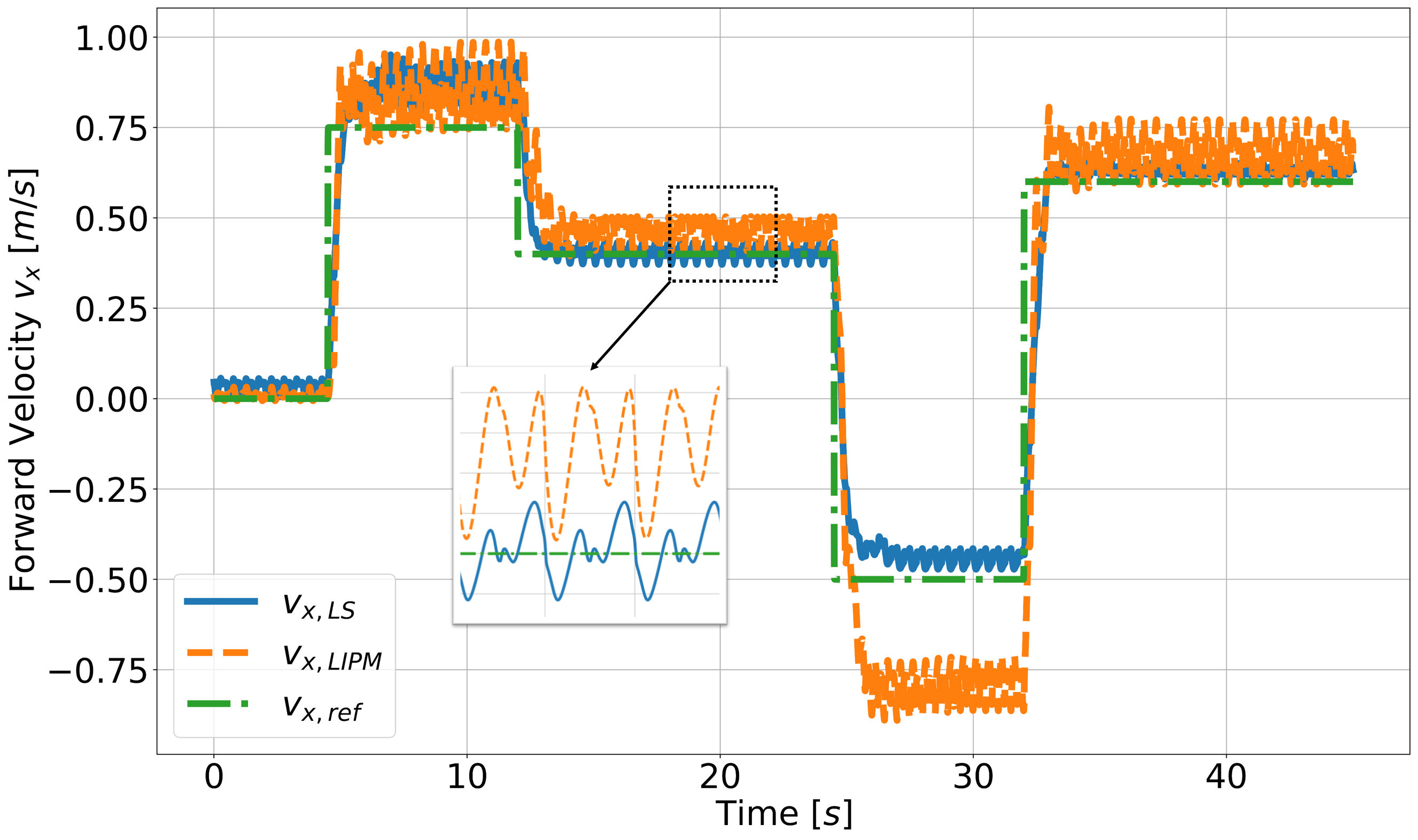}
\end{center}
\caption{Velocity tracking performance on flat terrain for the probosed method at a step duration of 0.25s. The LS-based method outperforms the LIPM-based approach.}
\label{TrackVelocity}
\end{figure}

We compare the performance of the investigated methods in tracking a predefined forward velocity profile while walking on flat terrain, as shown in Fig.~\ref{TrackVelocity}. Table~\ref{velocity_tracking_table}  shows the mean absolute error and the standard deviation of the forward base velocity at a reference velocity of 0.5 m/s, measured over 100 steps with a step duration of 0.25s. Our method achieves a mean absolute error (MAE) of 0.0136 m/s, which is about 80\% lower than that of the other method, demonstrating significantly improved accuracy. It also has the lowest standard deviation (0.0141 m/s), indicating more consistent performance. 

\begin{table}[t]
\caption{Velocity tracking performance}
\label{velocity_tracking_table}
\begin{center}
\begin{tabular}{lll}
\multicolumn{1}{c}{\bf Metric} & \multicolumn{1}{c}{\bf LIPM-based} & \multicolumn{1}{c}{\bf LS-based} \\ \hline \\
Mean absolute error [m/s] & 0.0767 & 0.0136 \\ \\
Standard deviation [m/s] & 0.0540 & 0.0141 \\
\end{tabular}
\end{center}
\end{table}

\subsubsection{Desired step location tracking}
Table~\ref{step_location_table} compares RL controllers in tracking the desired step location. The error is measured as the mean absolute error (MAE) between the planned and actual step locations while the robot walks at an average speed of \(0.6 \, \text{m/s}\) over 20 steps.

\begin{table}[t]
\caption{Step location tracking performance}
\label{step_location_table}
\begin{center}
\begin{tabular}{lll}
\multicolumn{1}{c}{\bf Metric} & \multicolumn{1}{c}{\bf LIPM-based} & \multicolumn{1}{c}{\bf LS-based} \\ \hline \\
Mean absolute error [m] & 0.003 & 0.002 \\
\end{tabular}
\end{center}
\end{table}

\subsubsection{Performance at different step duration (walking frequency)}
We trained our controllers to work with different step durations (or step frequencies). In general, the robot successfully adjusted its step duration within the defined range of \([0.2, 0.4]\)s. However, its ability to follow velocity commands decreased as the step duration increased. Nevertheless, the robot remained stable and was able to walk without falling over the entire range.

\subsubsection{Energy Efficiency}
The energy efficiency of the control systems was evaluated based on the mechanical and electrical costs of transportation \citep{torricelli2015benchmarking}. We simulated the robot walking on flat ground at an average speed of \(0.6\,\text{m/s}\) over 100 steps. The results are presented in Table~\ref{energy_table}. The LS-based approach outperformed the LIPM-based method, even though the latter relies on robot dynamics to determine optimal positions.

\begin{table}[t]
\caption{Cost of transport across the studied methods}
\label{energy_table}
\begin{center}
\begin{tabular}{lll}
\multicolumn{1}{c}{\bf Metric} & \multicolumn{1}{c}{\bf LIPM-based} & \multicolumn{1}{c}{\bf LS-based} \\ \hline \\
Electrical Cost ($C_{et}$) [J/kg·m] & 0.41 & 0.35 \\ \\
Mechanical Cost ($C_{mt}$) [J/kg·m] & 0.20 & 0.19 \\
\end{tabular}
\end{center}
\end{table}

\subsubsection{Robustness}
We evaluated the robustness of the controllers by applying external linear pushes to the robot's base and recording its response (fall or recover). Each test consisted of 500 samples, with maximum impulse values of \(7 \, \text{N·s}\) and \(5 \, \text{N·s}\). The robot was instructed to maintain zero velocity throughout the tests.

\begin{figure*}[!t]
\begin{center}
\includegraphics[width=\textwidth]{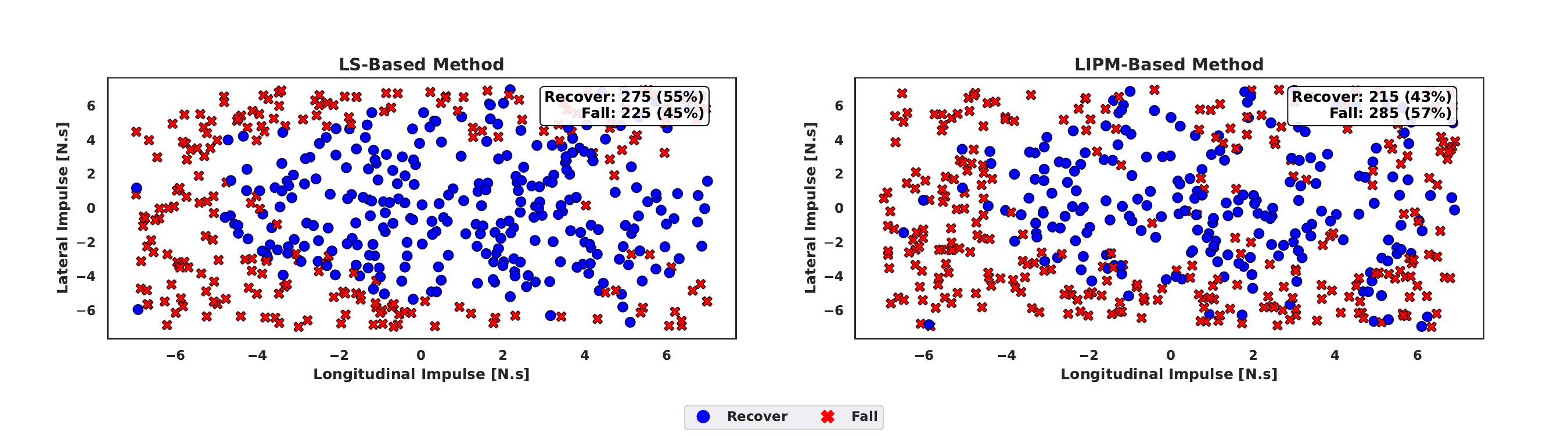}
\end{center}
\caption{Robot’s response to a push at zero commanded velocity. The LS-based method was able to handle more pushes, with a specific force applied, than the other method, demonstrating better stability under disturbances. The total number of tested points is 500. Below each subplot, the total number of recovered and fallen points is indicated.}
\label{pushrecovery}
\end{figure*}

The recovery success rates are shown in Table~\ref{push_recovery_table}. Fig.~\ref{pushrecovery} shows the distribution of coordinates where the robot either falls or successfully recovers for \(I_{\text{max}} = 7 \, \text{N·s}\). The results show that the LS-based method outperforms the other approaches.

\begin{table}[t]
\caption{Push recovery success rates}
\label{push_recovery_table}
\begin{center}
\begin{tabular}{lll}
\multicolumn{1}{c}{\bf Impulse Magnitude} & \multicolumn{1}{c}{\bf LIPM-based} & \multicolumn{1}{c}{\bf LS-based} \\ \hline \\
7 [N·s] & 0.43 & 0.55 \\ \\
5 [N·s] & 0.68 & 0.81 \\
\end{tabular}
\end{center}
\end{table}

 \subsubsection{Walking on rough terrain}
\begin{figure}[ht]
\begin{center}
\includegraphics[width=0.8\columnwidth]{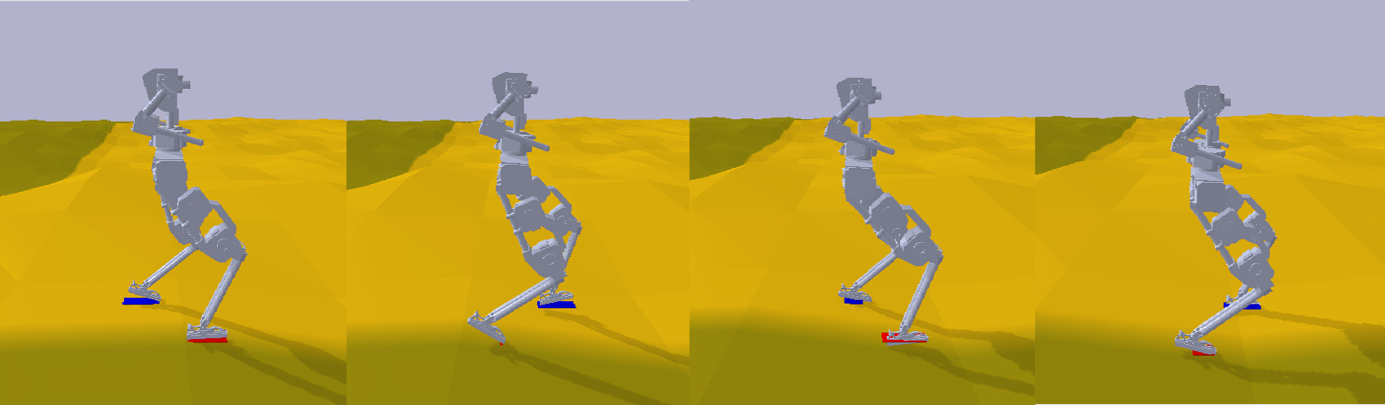}
\end{center}
\caption{Blind walking on rough terrain using the proposed LS-based approach in the PyBullet environment.}
\label{roughterrainsim}
\end{figure}

\begin{figure}[ht]
\begin{center}
\includegraphics[width=0.7\columnwidth]{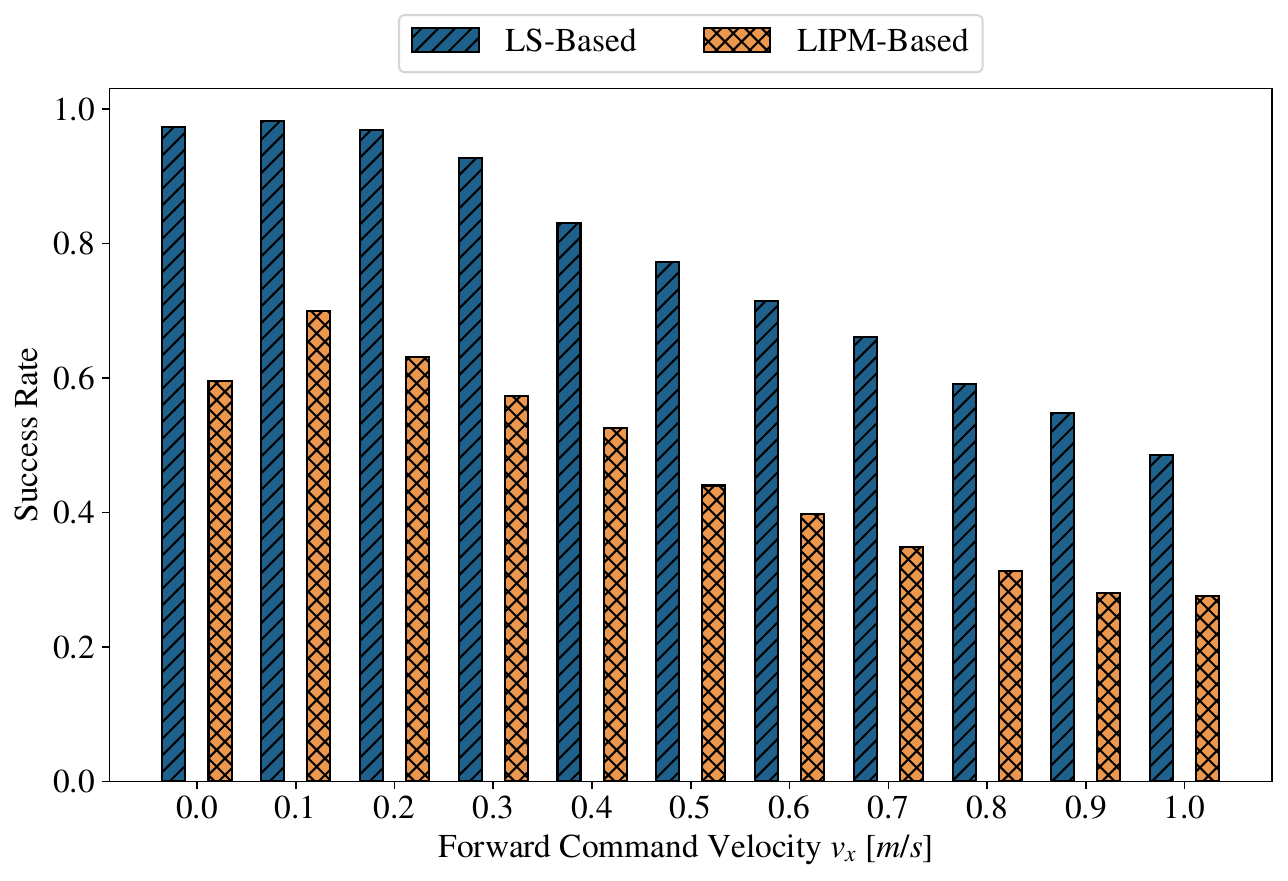}
\end{center}
\caption{Comparison of success rates for walking on rough terrain. The LS-based method demonstrated excellent performance in blind walking on rough terrain, exhibiting approximately 60\% greater stability in trials compared to the other method.}
\label{roughsuccessrate}
\end{figure}

To evaluate the adaptability of our policy to previously unseen and uneven terrain, we conducted experiments on rough terrain (Fig.~\ref{roughterrainsim}). The evaluation metric was the robot's ability to maintain a predefined forward velocity command for five seconds without falling. As shown in Fig.~\ref{roughsuccessrate}, our approach achieved a higher success rate than the other approach. Although we applied the modifications proposed in \citep{lee2024integrating} to improve the LIPM-based method, the simple heuristic approach exhibited approximately 60\% greater stability in trials compared to the LIPM-based method. In particular, our method worked in these experiments without incorporating any information about the height of the feet into the policy.

\subsubsection{Walking on a Terrain with Gaps}
To evaluate the effectiveness of the proposed method, we assessed the robot’s performance on a surface containing gaps, using the same policy trained on a flat surface in the IsaacGym simulation environment. Terrain perception was based on a height map, consistent with the conditions used during training for stair traversal. When a gap was detected in the region of the next planned step, the robot shifted the foot placement to the nearest available flat surface. A similar strategy was employed in \cite{lee2024integrating} to enable gap crossing in legged robots.

\begin{figure}[ht]
\centering
\includegraphics[width=0.80\columnwidth]{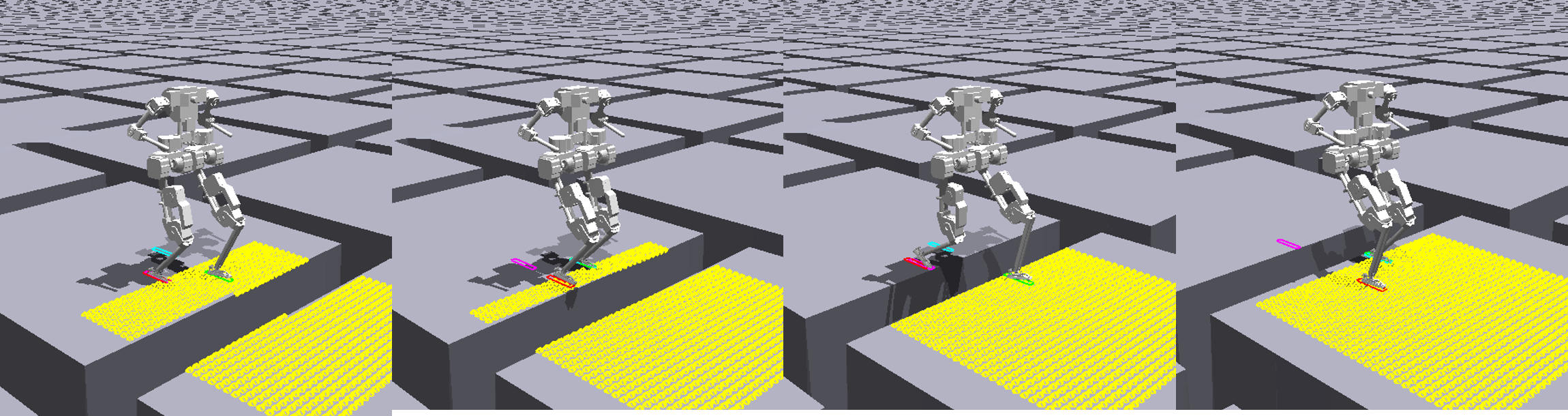}
\caption{Evaluation of the proposed approach for walking on terrain with gaps. The robot successfully crosses gaps with widths not exceeding the maximum step length (30 cm). Red — commanded position of the right foot; blue — commanded position of the left foot; yellow dots — terrain height map.}
\label{GapTerrain}
\end{figure}

Figure~\ref{GapTerrain} illustrates the robot successfully walking across a surface with gaps while following the generated step commands at a walking speed of 0.6 m/s. This demonstrates the capability of the proposed method to maintain stability and dynamically adjust foot placements when traversing discontinuous terrain.

\subsection{Discussion}

The proposed method, which integrates simple heuristic step planning into the learning architecture, has demonstrated high adaptability. Although the policy was trained on flat terrain, it successfully handles disruptions in surface structure. This generalization capability confirms the effectiveness of the approach for walking across diverse terrain types.

An analysis of the obtained results shows that the studied approaches deliver good performance despite their structural differences. This indicates that step-planning methods are a promising strategy for controlling bipedal walking robots, as they do not restrict the robot's ability to exploit the full dynamics of its body.

When following a target velocity profile, the methods demonstrated strong results. However, the LS approach achieved a mean absolute error of \(0.0136\,\text{m/s}\), which is approximately 80\% lower than that of the other method, indicating substantially higher velocity-tracking accuracy.

The robot was able to walk with various step frequencies. We believe that training the robot with different walking frequencies allowed it to exploit its dynamics more effectively. This is supported by the observation that walking with increased step duration resembled jogging rather than standard walking. Furthermore, under external disturbances, the robot's responses were similar to those observed when using longer step durations.

The LS-based method demonstrates superior energy efficiency compared to the LIPM-based approach. This improvement stems from its simpler controller structure, which grants the reinforcement learning policy greater freedom to explore optimal trajectories, ultimately leading to enhanced energy efficiency.

The LS method demonstrated an approximately 20\% higher disturbance rejection capability compared to the other method under an impulse force of \(5 \, \text{N·s}\). Moreover, it achieved excellent results in "blind" walking over uneven terrain, providing approximately 50\% higher stability in tests compared to the other method.

One of the most important findings of our study is that achieving high performance and robustness is possible even without relying on the dynamic model of the robot. The proposed heuristic method has shown that simple step planning can be sufficient to form an effective and generalizable control strategy. This confirms that model-based methods are not strictly necessary for providing guidance during learning.

\section{CONCLUSIONS}
\label{conclusions}

In this work, we proposed an approach for learning a policy that accurately tracks desired step locations. Our method uses a simple linear step planner controller, with the reinforcement learning (RL) policy ensuring stability. We compare its performance against alternative approach based on a reduced-order model. The results show that even a simple controller is sufficient to achieve the learning objective and excels in several key performance measures, such as velocity tracking, energy efficiency, and robustness. This comparison also addresses a critical question about the need for model-based methods to guide the RL policy training, and examines how the chosen method influences performance.

In future work, we will focus on the further development of the LS-based RL controller. Our goals are twofold: 1) To evaluate the performance of this controller on the real hardware of BRUCE. 2) As mentioned earlier, the controller in this study did not use any step height information. The next step is to train the LS-based RL controller to adapt to variations in ground level by adjusting the step height. This extension will serve as the basis for the development of an RL-based perspective controller that will replace the LS-based high-level controller and work in conjunction with the low-level policy to manage rough terrain with gaps and varying step heights. 

\acks{The authors would like to express their gratitude to all colleagues and reviewers whose comments and suggestions greatly improved this work.}

\bibliography{sample}

\end{document}